\pdfoutput=1
\documentclass[10pt,twocolumn]{article} 
\usepackage{simpleConference_alex}
\usepackage{times}
\usepackage{graphicx}
\usepackage{amssymb}
\usepackage{url,hyperref}
\usepackage{amsmath}
\usepackage{adjustbox,lipsum}
\newcommand*\samethanks[1][\value{footnote}]{\footnotemark[#1]}

\begin{document}

\title{\textbf{Keys to Accurate Feature Extraction Using Residual Spiking Neural Networks}}

\author{Alex~Vicente-Sola \thanks{Corresponding author: alex.vicente-sola@strath.ac.uk}~~\thanks{Neuromorphic Sensor Signal Processing Lab, Centre for Image and Signal Processing, Electrical and Electronic Engineering, University of Strathclyde, Glasgow, UK.}~,
        Davide~L.~Manna \samethanks~,
        Paul~Kirkland \samethanks~,
        Gaetano~Di~Caterina \samethanks~,
        
       and Trevor~Bihl

%\protect\\
\thanks{ Air Force Research Laboratory, Wright Patterson AFB, OH}% <-this % stops an unwanted space}
}

\maketitle
\thispagestyle{empty}

\begin{abstract}
Spiking neural networks (SNNs) have become an interesting alternative to conventional artificial neural networks (ANN) thanks to their temporal processing capabilities and energy efficient implementations in neuromorphic hardware. However the challenges involved in training SNNs have limited their performance in terms of accuracy and thus their applications. Improving learning algorithms and neural architectures for a more accurate feature extraction is therefore one of the current priorities in SNN research. In this paper we present a study on the key components of modern spiking architectures. We design a spiking version of the successful residual network architecture and provide an in-depth study on the possible implementations of spiking residual connections. This study shows how, depending on the use case, the optimal residual connection implementation may vary. Additionally, we empirically compare different techniques in image classification datasets taken from the best performing networks. Our results provide a state of the art guide to SNN design, which allows to make informed choices when trying to build the optimal visual feature extractor. Finally, our network outperforms previous SNN architectures in CIFAR-10 (94.14\%) and CIFAR-100 (74.65\%) datasets and matches the state of the art in DVS-CIFAR10 (72.98\%), with less parameters than the previous state of the art and without the need for ANN-SNN conversion. Code available at: \url{https://github.com/VicenteAlex/Spiking\_ResNet}
\end{abstract}

\section{Introduction}\label{sec:introduction}

Artificial Neural Networks (ANNs) have achieved in recent years unprecedented performances in many computer vision tasks. However, these artificial systems still cannot be compared to a real brain in terms of robustness, energy consumption or generalization capabilities. Therefore, as an attempt to imitate more of the valuable properties of the brain, artificial Spiking Neural Networks (SNNs) have been proposed as an alternative to conventional ANNs. SNNs closely replicate the functioning of biological neurons, allowing for sparse asynchronous computations and time-dependent neuronal functionality. The full potential of these properties is yet to be explored, but it has already been proved how substantial improvements in energy efficiency can be obtained by implementing SNNs in neuromorphic hardware \cite{davies2018loihi,cassidy2013cognitive}, bringing efficiency gains of up to 100 times less compared to standard ANNs in CPU/GPU hardware \cite{davies2021advancing}. Given the ever increasing network size and power demands of standar ANNs, such energy efficiency gains are of particular interest as they allow to reduce SWaP (Size, Weight, and Power) for energy efficiency operations \cite{anthony2020carbontracker}.

However, training SNNs is a more challenging task than training regular non-spiking networks. Non-spiking ANNs owe most of their success to the back-propagation of error (BP) algorithm \cite{Lecun2015}, but in the case of SNNs the spiking behaviour inside the neurons creates a non-differentiable function, hindering the application of BP. Moreover the time dependencies of the neuronal states add extra complexity to the credit assignment calculations. These drawbacks result, in most cases, in SNNs having a lower final accuracy than regular ANNs.

In order to overcome the aforementioned challenges, some approaches use conversion methods \cite{esser2016convolutional,Kugele2020,stockl2019recognizing}, where they train non-spiking ANNs and then approximate their computations using an SNN. Compared to directly training an SNN, these methods are not able to perform online learning, they lose temporal resolution, and in most cases they have higher latency and energy consumption. This is why improving directly trained SNNs is still a necessity.  

Direct training can be performed through bio-plausible unspervised methods such as  Spike-timing-dependent plasticity (STDP), but when ground truth is available for the task to solve, supervised learning through surrogate gradient BP \cite{neftci2019surrogate} is the best performing method. In this work we focus on the latter.

In order to improve the feature extraction process of SNNs in visual tasks, in this paper we present a study on the key components of modern spiking architectures and use the obtained conclusions to propose a novel and highly optimized SNN. Our results prove how directly training SNNs can already outperform conversion methods, allowing to exploit all the benefits of spiking computations without compromising accuracy. Additionally, the lessons learned from our experiments can also be valuable for those designing new SNN feature extractors in the future.

Specifically, the contributions of the paper are as follows. First, it presents an in-depth study on the possible implementations of spiking residual connections which highlights their properties in terms of accuracy, network activity, characteristics of their derivatives and implications of the computations in hardware requirements. This study introduces a novel residual connection for SNN which has been named the "Voltage to Voltage" connection and a revamped implementation of the "Spikes to Spikes" connection.

Then, it provides empirical results demonstrating the effects of different network design choices on the final accuracy. These include network size, batch normalization strategies, boosting methods, spike generation for frame based datasets, hyper-parameter optimization and fine-tuning. When designing an SNN, the conclusions drawn from these experiments allow to make optimal design choices maximizing the accuracy of the system.

Finally, a new spiking network is defined which achieves higher accuracy than the previous state of the art in CIFAR-10 and CIFAR-100, and matching it for DVS-CIFAR10 with many less parameters than previous methods.

Additionally, a study on the compromise between latency and accuracy is presented. Through the experiments performed in it we also obtain novel results demonstrating a relationship between the processing time and the optimal leakage factor for a leaky integrate-and-fire model.

\section{Related work}

As mentioned in the previous section, one limitation in implementing SNNs is the difficulty to train them.  Conventional gradient descent algorithms are not directly applicable given the intrinsic presence of non-differentiable spiking functions, as a result, different workaround strategies have been proposed. These strategies can be mainly categorized into two groups, ANN to SNN conversion methods and direct training methods. In this section, we overview the state of the art of this two approaches.

\subsection{Conversion methods}
In order to overcome the challenges in SNN training and to obtain the most accurate SNN systems, many works have adopted conversion approaches. These methods allow to bypass the training challenges of SNNs by training a non-spiking ANN and then transforming it to spiking format. This transformation reconstructs each of the neurons in the original network using spiking neurons, therefore the key challenge is to represent continuous activation values using the binary outputs of spiking neurons.

Most of these techniques are based on rate-based conversion \cite{esser2016convolutional, Kugele2020, han2020rmp, wang2020deep}, where the network is set up such that  the spiking frequency of the converted neuron is proportional to the activation value of the original one. These methods can only convert ANNs using the Rectified Linear Unit (ReLU) activation function.

In order to reduce the energy cost of these conversions, Temporal-SwitchCoding (TSC) \cite{han2020deep} was proposed, where the activation value is encoded in the latency of spiking rather than the frequency thus generating less spikes. On the other hand, methods such as \cite{deng2021optimal, wu2021progressive} focus on reducing the conversion error without the use of a large number of time-steps, which allows for competitive results without long simulation times.

Finally, ReLU networks can also be approximated using the method in \cite{severa2019training}, where a binary ANN is trained in order to approximate the original in just one time-step. The reported results are less accurate than state of the art SNN conversions, but they allow for a 1 step inference without temporal computations.

Alternatively, other approaches such as \cite{stockl2019recognizing, stockl2021optimized} can be applied to any type of network. The first one manages to do this by using circuits of neurons in order to approximate arbitrary functions. The second one does the same by using FS-neurons, a parametric neuron model that can be optimized to approximate any function.

Converted networks can be implemented in energy efficient neuromorphic hardware; however, forcing the SNN to imitate non-spiking computations makes it lose some of its properties. A converted network can not perform online learning and, because it approximates dense activation maps, it is prone to lose sparsity. Moreover, it has a lower temporal resolution, which is likely to cause under-performance when processing neuromorphic data as proved by \cite{Deng2020}.

\subsection{Direct training} 
Directly training the SNN without conversion allows one to exploit all its valuable properties; however, the challenge becomes then to successfully train it given that gradient descent based methods cannot be applied to non-differentiable spiking functions.  The most common strategy in state of the art methods is the use of surrogate gradients \cite{wu2018spatio,kim2020revisiting}, a method where the spiking function is used in the forward path, but when calculating its derivative in the backwards path, a continuous tractable function is used, which tries to approximate the behaviour of the real derivative.

Another option is to use a version of the SNN model that is directly differentiable. Some examples can be found in \cite{neftci2019surrogate}. We can find models using soft non-linearities \cite{huh2017gradient}, probabilistic models \cite{mostafa2018learning} or latency-based networks \cite{mostafa2017supervised}.

Alternatively, supervised learning can also be performed without the differentiation of the whole network. Some examples use local approaches with algorithms such as \cite{kaiser2020synaptic}, where the loss is computed locally in each neuron, or by using three factor learning rules \cite{kusmierz2017learning}.

Depending on the needs of the system, the optimal learning method might change, but when talking about final task accuracy, surrogate gradient BP is the best performing method so far. All the best SNN feature extractors consistently use this method, but the BP implementations and the surrogate functions they use vary between them.

Concerning the BP implementation, different variations can be found among the best performing networks. Some works such as \cite{kim2020revisiting} choose to simply unroll the network in time and use Back-propagation Through Time (BPTT). A slightly different implementation is found in \cite{lee2020enabling}, where the authors use a Spike-based BP algorithm which proposes a novel way of accounting for the leak factor of LIF neurons. Finally, there are also BP approaches where the input spikes are convolved with spike response kernels like in \cite{shrestha2018slayer}, which allows for convenient spike response implementations at the cost of saving more spike time-stamps in memory.

For the surrogate functions, there is no consensus either. We find triangle shape surrogates in \cite{kim2020revisiting}, rectangular shaped in \cite{zheng2020going}, and arc-tangent shaped in \cite{fang2021incorporating, fang2021deep}.

\subsection{SNN architectures}

Regarding the state of the art of SNN topologies, literature usually measures their feature extraction capabilities by assessing their image classification accuracy in public datasets. In the present day, among directly trained networks, the highest accuracies are reported for networks basing their topologies on VGG \cite{simonyan2014very} and ResNet  \cite{he2016deep} architectures.

In non-spiking deep learning, after the development of deep feed-forward networks such as VGG, the next big improvement came with the addition of residual connections. As demonstrated in \cite{he2016deep}, residual connections allowed to successfully train much deeper architectures, giving rise to a more accurate and efficient family of networks.

The reason for this improved performance is that residual connections help alleviate the problem of depth-induced accuracy degradation. Without residual connections, when increasing the depth of the network, the accuracy firstly saturates, but then it degrades rapidly. This is caused by the fact that extra layers increase the complexity of the problem to optimize, therefore it can get to a point where the benefit of adding extra layers does not compensate for the harm of increasing optimization difficulty.

The way residual networks solve this problem is by making the network easier to optimize. Given an input $x$ and the mapping function of a layer $F(x)$, the output of a layer with a residual connection will be:

\begin{equation}\label{eq.residual}
    H(x) = F(x) + x
\end{equation}

Then, the residual mapping $F(x) = H(x) - x$ should become easier to optimize than the original $F(x) = H(x)$. This is because an identity mapping $H(x) = x$ can be accomplished just by setting the weights in the layer to zero ($F(x) = 0$), allowing the network to easily ignore unnecessary layers, and therefore not degrading the result. Alternatively, when the optimal solution is not an identity mapping it might still be closer to it than to a zero mapping, making for a better initialization \cite{he2016deep}.

In order to port these benefits to SNN, Lee et al. \cite{lee2020enabling}, Zheng et al. \cite{zheng2020going} and Fang et al. \cite{fang2021deep} implement the first trainable spiking ResNets, managing to train deeper networks than VGGs and achieving competitive results. On the other hand, \cite{kim2020revisiting,fang2021incorporating} implement VGG-like architectures which are shallower, but larger in number of parameters. These non-residual feed-forward networks still outperform the aforementioned ResNets in many datasets (see Table \ref{tab.sota} in Section \ref{sec.results}).

\section{Methods}
\subsection{Spiking neuron model}

In order to perform their computations, SNNs simulate the behaviour of biological neurons by means of mathematical models. In this work we use the Leaky Integrate-and-Fire (LIF) model \cite{abbott1999lapicque}. Despite their simplicity, LIF neurons found great success in many state of the art systems.

The LIF model can be formulated as the differential equation seen in Eq. \ref{eq.lif}, where $U(t)$ is the membrane potential, $U_{rest}$ the resting potential, $\tau$ is the time constant and $I(t)$ is the input current. When the voltage $U(t)$ surpasses a set threshold $U_{th}$, the neuron emits a spike and the potential is reset by subtraction.

\begin{equation} \label{eq.lif}
    \tau \frac{du}{dt} = -\left( U(t) - U_{rest} \right) + RI(t)
\end{equation}

In order to easily program this behaviour in machine learning models, explicit iterative versions of this differential equation are used. Let $i$ be a post-synaptic neuron, $u_{i,t}$ is its membrane potential, $o_{i,t}$ its spiking activation and $\lambda$ the leak factor. The index $j$ belongs to the pre-synaptic neuron and the weights $w_{i,j}$ dictate the value of the synapses between neurons. Then, the iterative update of the neuron activation is calculated as follows:
\begin{equation}\label{eq.spike}
    o_{i,t} = g \left( \sum_{j} (w_{ij}o_{j,t})\ + \lambda \cdot u_{i,t-1} \right)
\end{equation}
where $g(x)$ is the thresholding function, which converts voltage to spikes:

\begin{equation}\label{eq.threshold}
g(x) = \begin{cases}
    & \text{1, if } x \geq  U_{th} \\ 
    & \text{0, if } x < U_{th} 
\end{cases}
\end{equation}

After spiking, a reset is performed by the subtraction $u_{i,t}^*=u_{i,t} - U_{th}$, where $u_{i,t}^*$ is the membrane potential after resetting.

\subsection{Spiking Residual Network}

With the objective of building the most accurate SNN feature extractor, our starting point is to implement a spiking residual network (S-ResNet).

The motivation to choose this architecture is that almost all the non-spiking state of the art ANNs make use of residual connections in order to allow for the training of very deep networks. On the contrary, in the SNN domain, the state of the art is still based in VGG-like architectures for datasets such as CIFAR-10, CIFAR-100 and DVS-CIFAR10. Therefore we define a new S-ResNet that will allow to outperform the previous state of the art and justify the use of residual connections also in the SNN domain.

\subsubsection{Implementation of a spiking residual connection}\label{sec.connections}
In order to design our S-ResNet, the first step is to define the implementation of the spiking residual connection. The skip connection in a non-spiking network just sums the activation value of a previous layer to the activation of the current one (Eq. \ref{eq.residual}), but when using spiking neurons this sum can be performed in several ways.

Given a multilayered feed-forward SNN of LIF neurons, the membrane state vector $u_{l,t}$ of a layer $l$ at time $t$ is given by Eq. \ref{eq.layer_memb}, where $o_{l,t}$ is the layer's spiking activation and $W_l$ the synaptic weight matrix. These spiking activations are obtained by means of the spiking function $g$ (Eq. \ref{eq.layer_act}).

    \begin{equation}\label{eq.layer_memb}
        u_{l,t} = W_{l-1}o_{l-1,t} + \lambda \cdot u_{l,t-1}
    \end{equation}
    
    \begin{equation}\label{eq.layer_act}
         o_{l,t} = g(u_{l,t})
    \end{equation}
    
Then, the residual information coming from a previous layer at position $l-n$ can be integrated to the current layer $l$ using one of the following strategies:

    \textbf{Spike output to membrane (S2M):} The spiking output of a previous layer $l-n$ feeds the membrane potential of the neurons in layer $l$. A set of synaptic weights $W^\prime_{l-n}$ will be needed to define the amount of voltage communicated by these spikes (Eq. \ref{eq.layer_s2m}). These weights will typically be a non-learnable parameter, then if $W^\prime_{l-n} = U_{th}$ the residual connection will implement an identity mapping when $W_{l-1}o_{l-1,t} + \lambda \cdot u_{l,t-1} = 0$. In any other case, the final activations are not guaranteed to be $o_{l,t}=o_{l-n,t}$.
    
    \begin{equation}\label{eq.layer_s2m}
        o_{l,t} = g(W_{l-1}o_{l-1,t} + \lambda \cdot u_{l,t-1} + W^\prime_{l-n}o_{l-n,t})
    \end{equation}
    
    Regarding its training through back-propagation, the properties of the residual connection can be observed in the network's derivative. Consider a generic residual block where the residual input $W^\prime_{l-n} o_{l-n,t}$ has $n=2$ (Eq. \ref{eq.layer_s2m}), skipping the intermediate layer $l-1$, and where $l-1$ has no residual input (Eq. \ref{eq.layer_s2m_2}). 
    
    \begin{equation}\label{eq.layer_s2m_2}
        o_{l-1,t} = g(W_{l-2}o_{l-2,t} + \lambda \cdot u_{l-1,t-1})
    \end{equation}
    
    Then, deriving Eq. \ref{eq.layer_s2m} with respect to $o_{l-2,t}$, we get:
    
    \begin{equation}\label{eq.de_s2m}
        \frac{\partial o_{l,t}}{\partial o_{l-2,t}}=\frac{\partial o_{l,t}}{\partial u_{l,t}} \frac{\partial u_{l,t}}{\partial o_{l-2,t}}=\frac{\partial o_{l,t}}{\partial u_{l,t}}( W_{l-1}\frac{\partial o_{l-1,t}}{\partial o_{l-2,t}} +
        W^\prime_{l-2})
\end{equation}
Eq. \ref{eq.de_s2m} shows how the residual connection adds an extra $W^\prime_{l-2}\frac{\partial o_{l,t}}{\partial u_{l,t}}$ term to the gradient, a term which is not influenced by the value of the learnable weights $W_{l-1}$, in contrast to $W_{l-1}\frac{\partial o_{l-1,t}}{\partial o_{l-2,t}}$. This is the reason why this residual connection will alleviate the vanishing gradient problem even when $W_{l-1}$ is arbitrarily small. Still, given that $\frac{\partial o_{l,t}}{\partial u_{l,t}}$ will be the derivative of the spiking function, the skip connection defined by this implementation will have its gradient scaled by the value of the surrogate function, which might be a concern depending on the setup.

The authors in \cite{fang2021deep} argue that the surrogate derivative $g^\prime$ of $g(u_{l,t})$ will typically not implement a function such that $g^\prime (W^\prime_{l-n} o_{l-n,t})=1$ when $o_{l-n,t} = 1$. Therefore scaling the derivative of the residual stream by this value could contribute to the vanishment or explosion of the gradient.
    
\begin{figure}[h]
\centering
\includegraphics[width=4in,trim={1.5in 5in 0in 0.8in},clip]{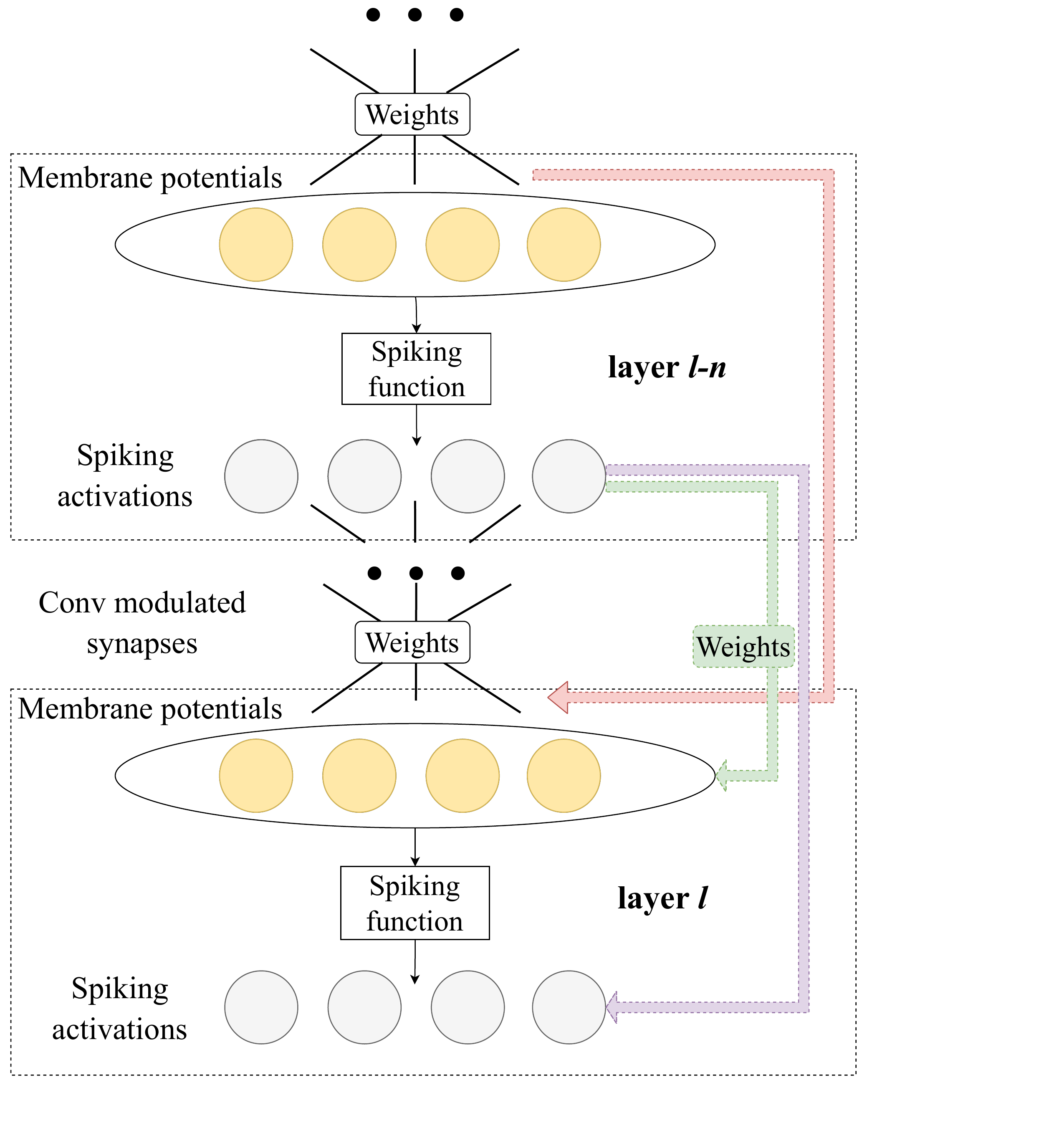}
\caption{The three possible residual connections in an SNN. In red: Membrane to membrane connection. Purple: Spike output to spike output. Green: Spike output to membrane. Note that the layers are displayed in one dimensional fashion for simplicity, but it is equivalent to a three dimensional convolutional map if the synapses are defined by a convolutional layer.}
\label{fig.residuals}
\end{figure}

    This kind of connection has previously been used in \cite{lee2020enabling} with $W^\prime_{l-n} = U_{th} = 1$ and in \cite{zheng2020going} weighted by their threshold-dependent batch normalization (potentially compromising the identity mapping). The S2M connection is represented in  Fig. \ref{fig.residuals} as the green connection.
    
    \textbf{Spike output to spike output (S2S):} The spiking output of a previous layer $l-n$ is added to the spiking output of layer $l$ (Eq. \ref{eq.layer_s2s}). If $o^\prime_{l,t} = 0$ this residual connection will successfully implement an identity mapping $o_{l,t} = o_{l-n,t}$.
        \begin{equation}\label{eq.layer_s2s}
        \begin{split}
        o^\prime_{l,t} = g(W_{l-1}o_{l-1,t} + \lambda \cdot u_{l,t-1})\\
        o_{l,t} = o^\prime_{l,t}  + o_{l-n,t}
        \end{split}
    \end{equation}
    Additionally, this implementation avoids applying the thresholding function to the residual path. Therefore, when using back-propagation, the contribution of the residual connection will be unaltered by the value of the surrogate function (Eq. \ref{eq.de_s22}).
    
    \begin{equation}\label{eq.de_s22}
        \frac{\partial o_{l,t}}{\partial o_{l-2,t}}=\frac{\partial o^\prime_{l,t}}{\partial u_{l,t}} \frac{\partial u_{l,t}}{\partial o_{l-2,t}} + \frac{\partial o_{l-2,t}}{\partial o_{l-2,t}} = \frac{\partial o^\prime_{l,t}}{\partial u_{l,t}} W_{l-1}\frac{\partial o_{l-1,t}}{\partial o_{l-2,t}} + 1
\end{equation}

    Regarding the information flow inside the SNN, this kind of connection has some implications that are worth noticing. It is implemented as an addition between activation maps, which is a different operation than adding voltages to a membrane and needs to be supported in the substrate implementing it (or else extra synapses will be needed). Moreover, it allows for the generation of non-binary activation maps, as the sum between activations could result in a value bigger than 1. In order to implement this, it will require to either sum activation maps and communicate non-binary values in the spike activation (as some neuromorphic hardware already supports \cite{orchard2021efficient}) or to avoid grouping spikes in one synapse by defining multiple individual connections such that:
    \begin{equation}
       o_{l,t} + (o_{l-n,t} + o_{l-m,t}) = o_{l,t} + o_{l-n,t} + o_{l-m,t} 
    \end{equation}
    
    Finally, in network topologies such as our S-ResNet (that we will define in the following section), we can find situations where the number of neurons $d_1$ in  $o^\prime_{l,t} \in \mathbb{N}^{d_{1}}$ is different than $d_2$ in $o_{l-n,t} \in \mathbb{N}^{d_{2}}$. As proposed in \cite{he2016deep}, we solve this by applying a 1$\times$1 convolution $f$ to $o_{l-n,t}$ such that $f:\mathbb{N}^{d_{2}} \rightarrow \mathbb{N}^{d_{1}}$. This is relevant for the S2S connection because, as seen in Eq. \ref{eq.layer_s2s_1x1}, by applying this convolution $o_{l-n,t}$ gets now multiplied by the learnable $W^{\prime\prime}$, which weights the activations transforming them into non-binary voltage values. The implications of these non-binary spiking activations are no different than that of the multiple spikes, it can be implemented as graded spikes in neuromorphic hardware or by defining extra synapses. The formulation for the later can be seen in Eq. \ref{eq.layer_s2s_1x1_u}, where the contribution of $o^\prime_{l,t}$ and $o_{l-n,t}$ to the membrane $u_{l+1,t}$ is split as two different incoming connections.
    
    \begin{equation}\label{eq.layer_s2s_1x1}
        o_{l,t} = o^\prime_{l,t}  + W^{\prime\prime}_{l-n}o_{l-n,t}
    \end{equation}
    
    \begin{equation}\label{eq.layer_s2s_1x1_u}
    \begin{split}
        u_{l+1,t} = W_{l}o_{l,t}  + \lambda \cdot u_{l+1,t-1}
        \\ = W_{l}o^\prime_{l,t} + W_{l}W^{\prime\prime}_{l-n}o_{l-n,t} + \lambda \cdot u_{l+1,t-1}
    \end{split}
    \end{equation}
    
    This kind of connection has been used in \cite{fang2021deep}. Its implementation is the same than the one in this work for maps at the same resolution, but it differs in the downsample paths. Differently from our proposal, a spiking neuron layer is added after the 1$\times$1 convolution. This was avoided in this work in order to eliminate the effect of the surrogate function in the derivatives of the residual path.
    
    The S2S connection is represented in  Fig. \ref{fig.residuals} as the purple arrow.
    
    \textbf{Voltage to voltage (V2V):} The previous two implementations created a residual mapping in the activation map. This residual mapping can also be enforced at the membrane potential level if a V2V connection is defined.
    
    Let the spiking input to a layer $l-n$ be $W_{l-n-1}o_{l-n-1}$ plus a residual input $r_{l-n,t}$. Then, in a V2V implementation, the input that feeds a layer $l-n$ will also become the residual input to the layer $l$ (Eq. \ref{eq.layer_V2V}). Like this, if $W_{l-1}o_{l-1,t} = 0$ and $u_{l,t-1} = u_{l-n,t-1}$ the residual will implement an identity mapping of the membrane potentials such that $u_{l,t} = u_{l-n,t}$. This will also cause $o_{l,t} = o_{l-n,t}$ if the thresholds of the two layers are the same.

    \begin{equation}\label{eq.res_V2V}
    r_{l,t} = W_{l-n-1}o_{l-n-1,t} + r_{l-n,t}
    \end{equation}
    
    \begin{equation}\label{eq.layer_V2V}
        o_{l,t} = g(W_{l-1}o_{l-1,t} + \lambda \cdot u_{l,t-1} + r_{l,t})
    \end{equation}
    Regarding the derivative of the network, deriving with respect to $o_{l-n-1,t}$ in the same setup as before ($n=2$) we get:
    
    \begin{equation}\label{eq.de_V2V}
\frac{\partial o_{l,t}}{\partial o_{l-3,t}}=\frac{\partial o_{l,t}}{\partial u_{l,t}} \frac{\partial u_{l,t}}{\partial o_{l-3,t}}=\frac{\partial o_{l,t}}{\partial u_{l,t}}( W_{l-1}\frac{\partial o_{l-1,t}}{\partial o_{l-3,t}} +
W_{l-3})
\end{equation}
    
    As it happened for the S2M, the derivative of the residual path will also depend on the surrogate function. Still, in the context of a hierarchical network, compared to an S2M implementation, the surrogate derivative will have less influence on this residual path, as $r_{l,t}$ is a function of $r_{l-n,t}$, which does not depend on $\frac{\partial o_{l-n-1,t}}{\partial u_{l-n-1,t}}$. In the case of the S2M implementation the residual is $r_{l,t}=W^\prime_{l-n}o_{l-n,t}$ which fully depends on $\frac{\partial o_{l-n,t}}{\partial u_{l-n,t}}$ adding an additional spiking function into the residual path with each residual block.
    
    Finally, notice that implementing the V2V connection will have the same effect in the information flow than S2S. This is caused by the dependency of Eq.\ref{eq.res_V2V} on $r_{l-n,t}$. In Eq.\ref{eq.unrav} we unravel this expression in order to show how the voltage sent by the residual connection $r_{l,t}$ is just a sum of post-synaptic potentials (PSP) from previous layers $W_{l-i\cdot n-1}o_{l-i\cdot n-1,t}$. Therefore, this can be implemented either by defining $(l/n)-1$ extra connections per each $r_{l,t}$ or by summing the PSPs together and then communicating the voltage value through graded spikes.

    \begin{equation}\label{eq.unrav}
        r_{l,t} = \sum_{i=1}^{(l/n)-1} W_{l-i\cdot n-1}o_{l-i\cdot n-1,t}
    \end{equation}
    
    The V2V connection is represented in  Fig. \ref{fig.residuals} as the red connection.\\

From an implementation point of view, this analysis showed how an S2M connection can be accomplished by a single conventional synapse while S2S and V2V require either to define multiple synapses or to perform a special kind of computation. This computation requires to sum spiking activations together for the S2S connection and to sum PSPs together in the case of V2V. Then the resulting value is transmitted to the membrane of the target neuron. With the neuromorphic hardware available in the present day, this could be implemented by an intermediate neuron which performs the sum and then transmits graded spikes.

In this work, we test the three approaches (section \ref{sec.experiments}) analysing their spiking activity (Fig. \ref{fig.activations}) and final accuracy (Table \ref{tab.connections}). We choose S2S for the final implementation, as it provides the most accurate results. This is consistent with the previous theoretical analysis, as S2S is the only solution avoiding spiking functions in the residual path.

\subsubsection{Network topology}\label{sec.topology}

With the residual connection implementation defined, the following choice to be made is the global network architecture. In the non-spiking domain it has already been proven how the original ResNet architecture \cite{he2016deep} outperforms feed-forward architectures without residuals; therefore, in order to test if the same principles apply to SNN, the obvious choice is to reuse the same topology.

Depending on the resolution and complexity of the dataset to target, the optimal architecture can vary; that is why in \cite{he2016deep} the architecture used for the ImageNet dataset and for CIFAR-10 are different. CIFAR images have a resolution of 32$\times$32, while the images are 224$\times$224 for ImageNet (after resizing), meaning that more downsampling operations will be needed in the second one in order to have a comparable receptive field. As we are targeting CIFAR-10, CIFAR-100 and DVS-CIFAR10, we will base our global network architecture on the smaller ResNet proposed for these datasets. The architecture is defined in \cite{he2016deep} in a table, such as Table \ref{tab.resnet_arch}.

\begin{table}[]\centering 
\caption{\label{tab.resnet_arch} Table defining the CNN architecture of the original ResNet proposed for the CIFAR datasets. The variable \(n\) allows to control the depth of the network.}
\begin{tabular}{c|c|c|c}
\hline
\textbf{output map size}                   & 32×32                        & 16x16 & 8x8 \\ \hline
\textbf{\# layers} & 1+2n & 2n    & 2n  \\ \hline
\textbf{\# filters}                        & 16                           & 32    & 64  \\ \hline
\end{tabular}

\end{table}

\begin{figure}[!t]
\centering
\includegraphics[width=2.7in,trim={3in 0.7in 3.2in 0.77in},clip]{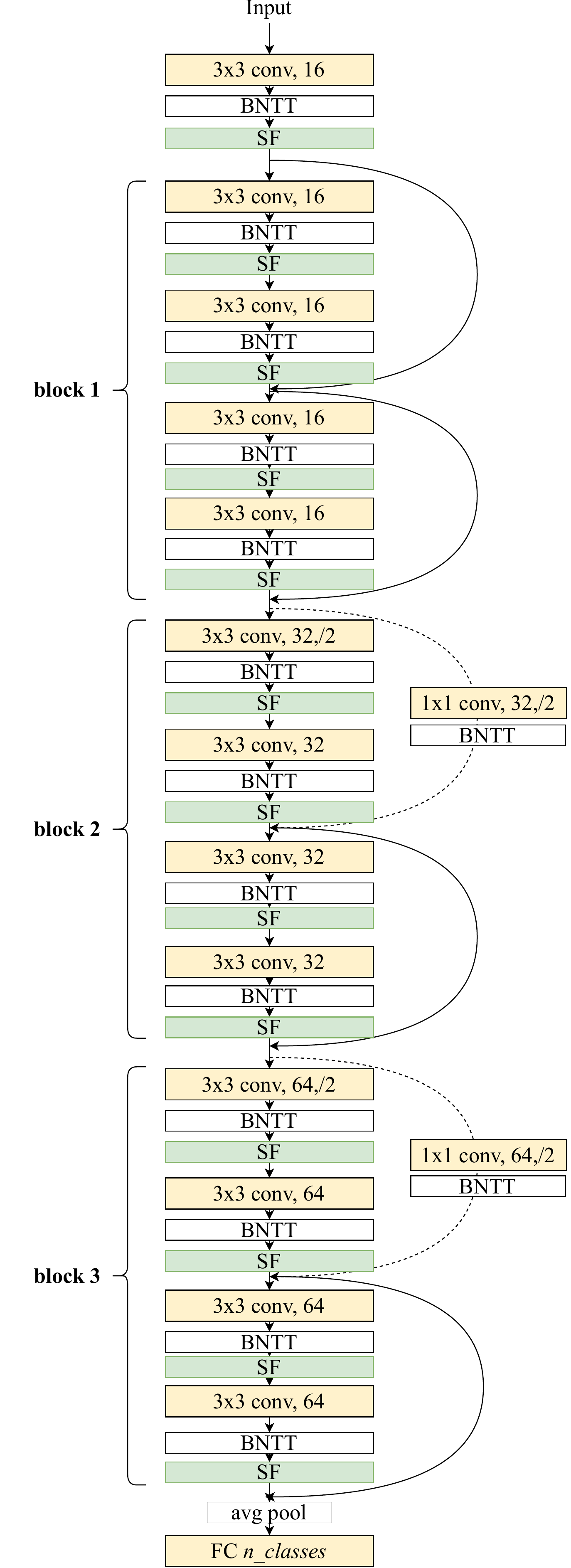}
\caption{Example architecture for an S-ResNet with \(n=2\) and 16 base filters. SF stands for spiking function. }
\label{fig.sresnet}
\end{figure}

Regarding the batch normalization (BN) layers in the architecture, regular BN can be used in an SNN, but improved performance has been reported by using Batch Normalization Through Time (BNTT) \cite{kim2020revisiting}, a time-varying BN that learns different statistics for each time-step. This is consistent with the studies performed in non-spiking RNNs, where works such as \cite{cooijmans2016recurrent} argue that the statistics of different time-steps can differ significantly. For that reason, in our final architecture we use BNTT. As further proof, Table \ref{tab.bn} in Section \ref{sec.experiments} demonstrates the performance gains of using BNTT compared to regular BN. A diagram of the final architecture can be found in Fig. \ref{fig.sresnet}.

To the best of our knowledge, this work is the first to implement the aforementioned architecture for SNN training. \cite{lee2020enabling, zheng2020going} implement alternative topologies with extra fully connected layers and larger amounts of channels in convolutional layers (see the difference in parameters in Fig. \ref{fig.cifar10} in Section \ref{sec.results}). The authors in \cite{fang2021deep} define their main network for ImageNet and reuse the original ResNet's topology for this dataset which is different from the CIFAR-10 one. Additionally, they propose a residual network targeting DVS-CIFAR10. Compared to ours, this network is wider and shallower (resulting in a larger parameter count), instead of strided convolution, it relies on max pooling for downsampling and it processes inputs of 128$\times$128 resolution. 

Apart from that, those three networks differ from ours in the normalization strategies, as they use time averaged statistics where we use BNTT, and also in the residual connection implementation.

\subsubsection{Boosting strategies}\label{sec.boosting}

Boosting techniques allow to combine the predictions of multiple weak classifiers to create a stronger one. Previous work in SNNs \cite{fang2021incorporating} has already applied simple versions of this strategy by converting the classification layer into a voting layer.

We tested the same approach as \cite{fang2021incorporating} and adapted the last fully-connected to have $10\times C$ neurons, where $C$ is the number of classes. Then an average pooling layer of kernel size 10 and stride 10 reduces the dimension back to the number of classes $C$. This process computes the score of each class as the average of 10 neuron states, which can be seen as a voting scheme for 10 different sub-networks.

In Section \ref{sec.experiments}, Tables \ref{tab.boost_cifar10} and \ref{tab.boost_cifar100} demonstrate the effects of adding the boosting layer. Some networks provided improved performance when using this strategy, while others did not, so we keep this layer only in those cases where it is beneficial. In our final results, only the CIFAR-10 network uses it.

\subsection{Training framework}

Our network is trained to perform image classification through supervised learning. In order to allow for this classification, the last neuron layer is defined with no leak and cannot spike. Then the voltage accumulated in the layer after $T$ time-steps divided by $T$ is considered the output value.

The output class scores are compared to the ground truth by means of a cross-entropy loss (Eq. \ref{eq.cross-entropy}), where $C$ is the number of classes, $u_{i,T}$ the voltage of neuron $i$ after the last time-step, and $y_i$ are the ground truth labels:

\begin{equation} \label{eq.cross-entropy}
    L = -\sum_{i}^{C}y_{i}log(\frac{e^{u_{i,T}}}{\sum_{j}^{C}e^{u_{j,T}}})
\end{equation}

With the loss defined, the weight updates for the learning process are calculated through BPTT.

The final voltage at each layer is dependent of the contribution of all previous time-steps, therefore the derivative of the loss function with respect to the network weights can be defined as the sum in Eq. \ref{eq.loss_weight}, for neurons in the output layer, and as the sum in Eq. \ref{eq.loss_weight_hidden} for neurons in the hidden layers.

\begin{equation}\label{eq.loss_weight}
    \frac{\partial L}{\partial w_{i,j}}=\sum^{T}_{t=1}\frac{\partial L}{\partial u_{t,i}} \frac{\partial u_{t,i}}{\partial p_{t,i}}\frac{\partial p_{t,i}}{\partial w_{i,j}}
\end{equation}

\begin{equation}\label{eq.loss_weight_hidden}
    \frac{\partial L}{\partial w_{i,j}}=\sum^{T}_{t=1}\frac{\partial L}{\partial o_{t,i}} \frac{\partial o_{t,i}}{\partial u_{t,i}} \frac{\partial u_{t,i}}{\partial p_{t,i}}\frac{\partial p_{t,i}}{\partial w_{i,j}}
\end{equation}

where $p_{i,t}$ is the current transmitted through the synapses after applying the weights:
\begin{equation}
    p_{i,t} = \sum_{j} w_{i,j}o_{j,t}
\end{equation}
Then, taking into account the temporal dependency of the membrane potential along with its dependency on input spikes, we obtain:

\begin{equation}
        \frac{\partial L}{\partial u_{t,i}}=\frac{\partial L}{\partial o_{t,i}} \frac{\partial o_{t,i}}{\partial u_{t,i}} + \frac{\partial L}{\partial u_{t+1,i}}\frac{\partial u_{t+1,i}}{\partial u_{t,i}}
\end{equation}

Notice that $\frac{\partial o_{t,i}}{\partial u_{t,i}}$ requires to compute the derivative of the thresholding function, which is non-differentiable. We solve this by using a triangle shaped surrogate gradient. As in \cite{kim2020revisiting}, we set $\alpha=0.3$.

\begin{equation}
    \frac{\partial o_{t,i}}{\partial u_{t,i}} = \alpha \max \{0, 1-\vert u_{t,i}\vert\}
\end{equation}

In practice this can be easily implemented using auto-differentiation tools such as Pytorch \cite{paszke2019pytorch}.

\subsection{Input preprocessing}\label{sec.preprocess}

\textbf{Frame-based datasets:} 
Frame-based images need to be encoded into spikes in order for an SNN to process them. Works like \cite{kim2020revisiting} use a Poisson spike generation process which transforms the image frame into a sequence of spikes. Other works  \cite{fang2021incorporating, zheng2020going} feed the unprocessed frame to the first SNN layer, making the pixel intensity equivalent to a constant input voltage for the first neurons.

The latter allows for better results, as all of the information is presented at each time-step, while the former will require many steps to represent all of the information and will add variability to the data. Still, we believe using a spike generation process is a better representation of a scenario where the input data is spiking information (such as the data coming from event cameras), so choosing one method or another should depend on the objective of the simulation. Therefore in this work we use both approaches in order to compare results. Our best performing networks are trained without Poisson encoder in order to maximize accuracy. Additionally, images are always normalized with respect to the statistics of the dataset.

\textbf{Neuromorphic datasets:}
Data produced by neuromorphic cameras represent the changes in the scene, and these are often presented in event format. An event is a discrete package of information indicating location, time-stamp and polarity (i.e. change in brightness).

We use the events to build frames containing spiking activations. Such frames have two channels, one for positive polarity and one for negative, and they accumulate all events occurring in a time window. The size of the time window is defined by the amount of time-steps we want to have for each sequence. We implement this process using the SpikingJelly library \cite{fang2020other}. 

\textbf{Data augmentation:}
Frame based datasets were augmented using random horizontal flips and random crops.

\subsection{Hyper-parameters}

The performance of the proposed network depends on certain hyper-parameters, such as the leak factor of the membrane, the number of time-steps or the learning rate for training. The optimal value of these parameters varies depending on the architecture of the network, the training procedure and the task at hand. That is why in order to properly asses how useful an architecture or a training method is, we first need to find its optimal hyper-parameter setup.

We address this challenge by using BOHB \cite{falkner2018bohb}, a hyper-parameter optimization technique that combines Bayesian Optimization (BO) and Hyperband (HB), a multi-armed bandit strategy. Using this method, we optimize the hyper-parameters for S-ResNet38 in the CIFAR-100 dataset. The learning rate for this training is divided by 10 at 70\%, 80\% and 90\% of the training process. The resulting hyper-parameters are also used for the rest of networks and datasets, as with the hardware available we could not afford to run an individual search per setup.

The best performing parameters are: \textit{leak = 0.874, time-steps = 50, learning rate = 0.0268} for a batch size of 21.

Notice that the target of the search was only to optimize accuracy, therefore the amount of time-steps tends to be maximized as it has a  monotonically non-decreasing relationship with the accuracy. Section \ref{sec.timesteps} demonstrates the effects of reducing the amount of time-steps.

\section{Experiments}\label{sec.experiments}
In order to maximize the accuracy of our method, we conducted a search for the key components in state of the art architectures that allow for improved performance. In this section we present empirical results obtained from testing these components in our networks. The results from these comparisons allow us to compose a network which outscores previous approaches in multiple datasets.

\textbf{Residual connection implementation:}
In section \ref{sec.connections} three ways of implementing residual connections in SNN were defined. We tested the performance of S-ResNet38 with each one of them (Table \ref{tab.connections}). The highest accuracy is obtained by the S2S connection. This result is consistent with our theoretical analysis, as the residual path in S2S does not go through spiking functions, therefore it allows a better flow of the gradient during back-propagation. Still, the performance of the V2V implementation is very close. On the other hand, the S2M implementation has a substantially lower accuracy. This decrease in accuracy could potentially be attenuated with further hyper-parameter search and improved optimization, but we hypothesize that such setup is more difficult to find due to the less convenient gradient properties of S2M.

\begin{figure*}[]
\centering
\includegraphics[width=7in]{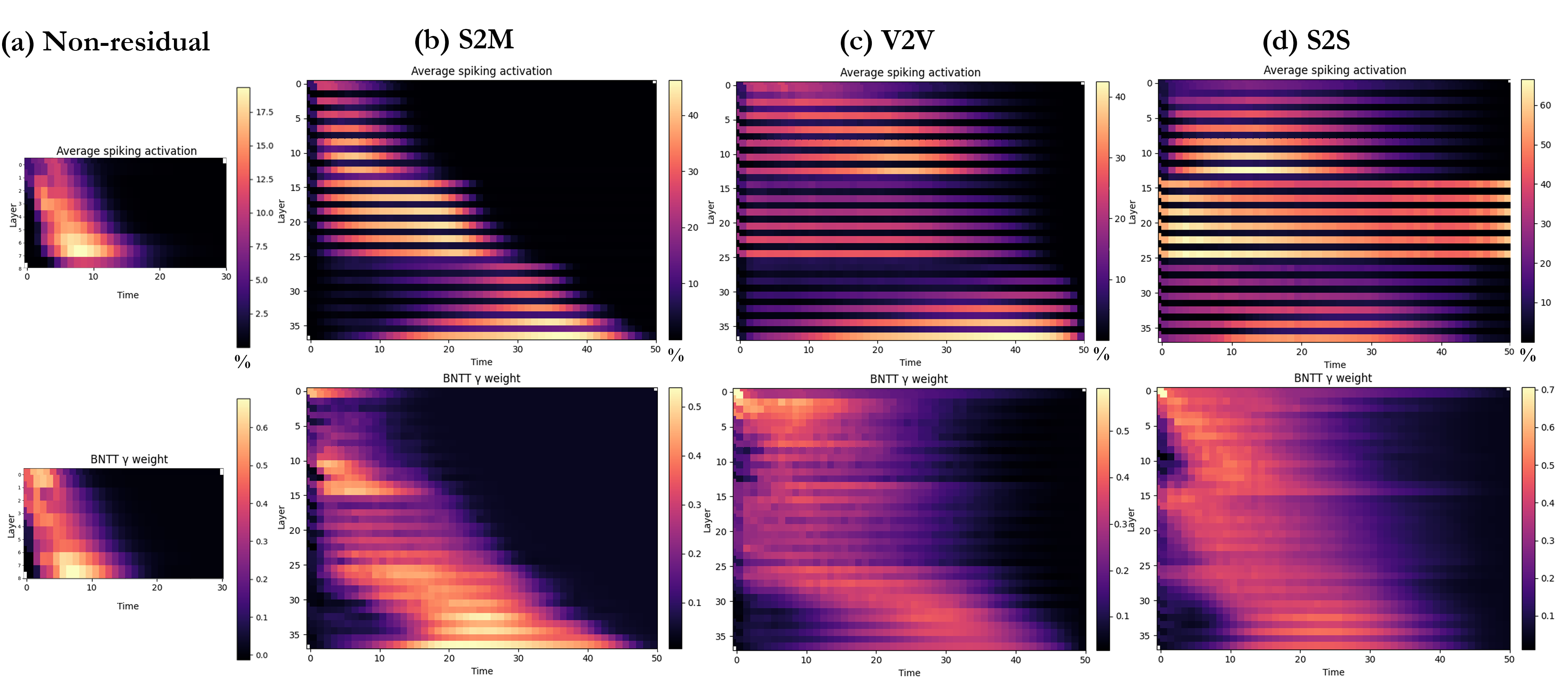}
\caption{\label{fig.spikes}Average activation maps of different networks in the CIFAR-100 test set (first row). The values represents the percentage of neurons active for each convolutional layer at each time. The second row displays the average value (over channels) of the learnable BNTT weight $\gamma$ per each layer and time-step. Column (a) uses a non-residual VGG11 architecture, (b),(c),(d) use our S-ResNet38 with 32 base filters.}
\label{fig.activations}
\end{figure*}

Apart from that, by adding any of these three residual connections, the network is expected to propagate more spikes to deeper layers. In order to analyse this effect, we averaged the spiking activity of the networks across the test set of the CIFAR-100 dataset (Fig. \ref{fig.spikes}). We also display the spiking activation obtained with a non-residual network (spiking VGG11) for comparison.

Before starting the comparison, it is important to realise the effect of BNTT in the spiking activation. As observed by \cite{kim2020revisiting}, by allowing to learn a different learnable weight $\gamma$ per time-step, the network is allowed to scale the activation of each layer depending on the time-step. Because of this, it tends to localise the spiking activity of each layer in a certain time range. The value of this weight for each network is visualized in the second row of Fig.\ref{fig.spikes}.

When looking at the S-ResNet networks, we observe how there are more layers active at each time-step, as the spiking connections propagate activations to deeper layers bypassing the BNTT weighing. The effect of BNTT is more noticeable in the S2M implementation and less in V2V and S2S. Still, all of them learn a time-dependant weight distribution, indicating that, according to back-propagation, that is the optimal solution for image classification. 

Apart from that, S-ResNet activity maps show a characteristic striped pattern. This is caused by how the residual connections always skip one layer, connecting only even-numbered layers (as defined in \cite{he2016deep}).

Finally, the more abrupt changes in activation percentage localized in layer 14 and 26 are caused by the resolution change, which changes the number of total neurons in the layer and makes the residual connection go through a 1$\times$1 convolution.

Overall the contribution of the residual connections behaves as expected. It propagates the spiking activations to deeper layers, which allows the back-propagation algorithm to successfully train deeper architectures. Additionally we see how the spiking activity is higher for S2S implementations compared to V2V or S2M, as the "multiple spikes" behaviour favours sending higher amounts of voltage between layers. This can be relevant for applications which are sensible to the volume of spiking activity. In those tasks, the optimal choice for the residual implementation can vary, as there is a compromise between accuracy and volume of spikes.

In cases where a lower network activation is needed V2V poses an efficient alternative to S2S with a very similar accuracy. Regarding their implementation, S2S and V2V require to define extra synapses per residual connection or to implement spike/PSP sum, therefore, S2M is the most suitable option for applications which want to avoid this.

\begin{table}[]
\centering
\caption{\label{tab.connections}Image classification test performance on CIFAR-10 and CIFAR-100. S-Resnet38 stands for the architecture defined in Section \ref{sec.topology} with \(n=6\) and 32 base filters, trained for 70 epochs.}
\begin{adjustbox}{max width=0.5\textwidth}
\begin{tabular}{|l|c|c|}
\hline
\textbf{Residual connection}   & \textbf{CIFAR-10 Accuracy}    & \textbf{CIFAR-100 Accuracy} \\ \hline
 
S-ResNet38 S2M      & 89.27 \% & 68.64 \%            \\ \hline
S-ResNet38 S2S & 94.01 \%  & 74.54 \%            \\ \hline
S-ResNet38 V2V     & 93.83 \% & 73.79 \%            \\ \hline
\end{tabular}
\end{adjustbox}

\end{table}

\textbf{Network depth:} The residual connections in S-ResNet allow to increase the depth of the network without the concern of catastrophic accuracy degradation. As expected, this allows us to train very deep architectures. Table \ref{tab.depth} presents the classification accuracy in CIFAR-10 achieved by the S-ResNet with different depths and the same training hyper-parameters. The results shows how the accuracy grows from 20 to 38 layers, but stays roughly the same from 38 to 44.

Given these results, for the rest of our experiments we choose S-Resnet38 as the default network. Still, the optimal depth of the network changes depending on the dataset and task to solve, therefore we encourage those researchers looking for optimal performance to tune this parameter for their specific task.

\textbf{Spike generation for frame based datasets:}
As mentioned in Section \ref{sec.preprocess}, when working with frame based datasets, we tested two different methods for the spike encoding process. One consists in transforming the intensity values into spikes by means of a Poisson spike generation process. The other consists in transforming them by means of the first convolutional layer (i.e. feeding the raw image to the network). 

As expected, the results in Table \ref{tab.poisson} show how encoding by means of the first convolutional layer gives a better result than generating spikes as a Poisson process. In order to maximize accuracy, for all of our experiments we use the encoding by convolution approach.

\textbf{Batch normalization strategies:}
We compare performances using time-dependent BN statistics versus time averaged statistics. Table \ref{tab.bn} shows how BNTT outperforms regular BN for the same network.

\textbf{Boosting layer:}
As introduced in Section \ref{sec.boosting}, a simple boosting layer can improve the accuracy of the system in some cases. Tables \ref{tab.boost_cifar100} and \ref{tab.boost_cifar10} show the effect of this component in the accuracy of our networks. In the CIFAR-10 datasets the accuracy is improved by using this technique, while in the CIFAR-100 one, where we have more classes, increasing the size of the last fully connected in order to perform boosting ends up being detrimental.

%It can be seen how in CIFAR-100, where we have more classes, our network does not have enough feature channels for the boosting layer to improve results. When the architecture is made wider the benefits of the voting strategy become noticeable again.

\textbf{Parametric Leaky Integrate-and-Fire:}
The authors in \cite{fang2021incorporating} propose to learn the leak coefficient of the LIF neurons directly through back-propagation as another parameter of the network. By doing this they can also afford to learn a different leak value for each layer. They call this method the Parametric Leaky integrate-and-fire (PLIF) neuron. Table \ref{tab.plif} shows our results after training S-ResNet38 with PLIF and with a single leak coefficient learned through hyper-parameter search.

We do not achieve our best results using the PLIF neuron; still, we believe this strategy is a very efficient way of finding this hyper-parameter. For this reason, we test it again for the search of a shared leak value instead of calculating a different one per layer. Table \ref{tab.leak} shows the difference between the leak value found through hyper-parameter search and the one found by back-propagation. It is interesting to see how the two values differ by a considerable amount, having the one found by back-propagation a slower leakage than the one found through the BOHB method.

Still, both values perform well when the network adapts its weights to work with them. The performance comparison between them can be found in Table \ref{tab.single_plif}, where we compare our network trained with the BOHB optimized value to an identical network which learned the shared leak value through PLIF.

\textbf{Extra training data:} 
In the deep learning domain, most state of the art performances in computer vision are achieved by means of fine tuning. This strategy consists in taking a network that has already been trained in a different dataset and then training it further for the task at hand. In the visual domain this strategy works well, as visual data has many transferable features.

We test this strategy by pre-training our networks with CIFAR-100 and then fine-tunning for DVS-CIFAR10 and CIFAR-10. The results are presented in Table \ref{tab.pretrain_dvscifar10} and Table \ref{tab.pretrain_cifar10}. We obtain higher accuracy results in all cases but for the larger S-ResNet in CIFAR-10. Moreover these trainings converge faster, making it a great solution for any further work building on top of these feature extractors. In our public code, users can find our pre-trained weights so that they can perform fine-tunning in any future system building from this one.

\textbf{DVS-CIFAR10 image resolution:} The event streams found in the DVS-CIFAR10 dataset were generated by recording 10,000 images from the original CIFAR10 dataset with a DVS camera while applying a repeated closed-loop smooth movement \cite{li2017cifar10}. Despite the resolution of CIFAR-10 being 32$\times$32, the DVS camera resolution was 128$\times$128 and therefore the resulting event maps have also a 128$\times$128 resolution. As our S-ResNet architecture is optimized for inputs of size 32$\times$32, in our previous experiments we downsampled the DVS-CIFAR10 dataset to that resolution.

In most datasets, downsampling the input causes information loss and therefore accuracy degradation. In order to test if this applies to the unique case of DVS-CIFAR10, we tested the performance using 64$\times$64 and 128$\times$128 resolution as input. We adapt the architecture of the network for the new input sizes by adding, in the case of 64$\times$64 a stride=2 in the first convolution (c32k3s2), and in the case of 128$\times$128 a stride=2 and kernel=5x5 in the first convolution (c32k5s2) followed by a Max Pooling of stride=2 and kernel=2 (MPk2s2).

Table \ref{tab.dvs_res} presents the test results with the three resolutions. It can be seen how the best performance is obtained when using a 64$\times$64. We do not obtain any improvement by using the full 128$\times$128 resolution. Our best architecture for full resolution uses a bigger kernel and max pooling, similarly to how \cite{he2016deep} handles the bigger ImageNet frames. We hypothesize that this setup does not bring improved performance because the down-scaled 64$\times$64 events already contain the necessary information and therefore the bigger 128$\times$128 network just brings unnecessary complexity.

\begin{table}[]
\centering
\caption{\label{tab.depth}Image classification test performance on CIFAR-10. S-Resnet stands for the architecture defined in Section \ref{sec.topology} with 16 base filters, trained for 70 epochs.}
\begin{adjustbox}{max width=0.5\textwidth}

\begin{tabular}{|l|c|}
\hline
\textbf{Network}   & \textbf{CIFAR-10 Accuracy}  \\ \hline
 
S-ResNet20      & 90.89 \%             \\ \hline
S-ResNet38 & 91.97 \%           \\ \hline
S-ResNet44     & 91.96 \% \\ \hline
\end{tabular}
\end{adjustbox}

\end{table}

\begin{table}[]
\centering
\caption{\label{tab.poisson}Image classification test performance on CIFAR-100. Except for the spike generation process, both architectures and training procedures are identical. Trained for 100 epoch.}
\begin{adjustbox}{max width=0.5\textwidth}

\begin{tabular}{|l|c|}
\hline
\textbf{Network}       & \textbf{CIFAR-100 Accuracy} \\ \hline
 
S-ResNet38 Poisson spike generation      & 64.96 \%            \\ \hline
S-ResNet38 Raw image & 69.03 \%            \\ \hline
\end{tabular}
\end{adjustbox}

\end{table}

\begin{table}[]
\caption{\label{tab.bn}Image classification test performance on CIFAR-100. Except for the batch normalization module, both architectures and training procedures are identical. S-ResNet stands for the architecture defined in Section \ref{sec.topology} with \(n=6\) and 32 base filters. Trained for 70 epochs.}
\centering
\begin{adjustbox}{max width=0.5\textwidth}

\begin{tabular}{|l|c|}
\hline
\textbf{Network}       &\textbf{CIFAR-100 Accuracy} \\ \hline
 
S-ResNet38 BNTT             & 74.54 \%            \\ \hline
S-ResNet38 BN time averaged & 70.82 \%            \\ \hline
\end{tabular}
\end{adjustbox}

\end{table}

\begin{table}[]
\centering
\caption{\label{tab.boost_cifar100}Image classification test performance on CIFAR-100. S-Resnet38 stands for the architecture defined in Section \ref{sec.topology} with \(n=6\) and 16 base filters. Wider architectures use 32 base filters and "boosting" indicates the use of a boosting layer (Section \ref{sec.boosting}). Wider architectures trained for 70 epochs, regular architectures trained for 200 epochs.}
\begin{adjustbox}{max width=0.5\textwidth}

\begin{tabular}{|l|c|c|}
\hline
\textbf{Network}   & \textbf{Parameters}    & \textbf{CIFAR-100 Accuracy} \\ \hline
 
S-ResNet38      &639,760 & 68.71 \%            \\ \hline
S-ResNet38 + boosting & 697,360   & 64.60 \%            \\ \hline
S-ResNet38 wider     & 2,399,776& 74.46 \%            \\ \hline
S-ResNet38 wider + boosting &2,514,976 & 73.21 \%            \\ \hline
\end{tabular}
\end{adjustbox}
\end{table}

\begin{table}[]
\centering
\caption{\label{tab.boost_cifar10}Image classification test performance on CIFAR-10. S-Resnet38 stands for the architecture defined in Section \ref{sec.topology} with \(n=6\) and 16 base filters. Wider architectures use 32 base filters and "boosting" indicates the use of a boosting layer (Section \ref{sec.boosting}). Wider architectures trained for 70 epochs, regular architectures trained for 200 epochs.}
\begin{adjustbox}{max width=0.5\textwidth}

\begin{tabular}{|l|c|c|}
\hline
\textbf{Network}   &\textbf{Parameters}    & \textbf{CIFAR-10 Accuracy} \\ \hline
 
S-ResNet38      & 634,000 & 91.97 \%            \\ \hline
S-ResNet38 + boosting & 639,760   & 92.00 \%            \\ \hline
S-ResNet38 wider &2,388,256 & 92.66 \%            \\ \hline
S-ResNet38 wider + boosting &2,399,776 & 93.77 \%            \\ \hline
\end{tabular}
\end{adjustbox}
\end{table}

\begin{table}[]
\centering
\caption{\label{tab.plif}Image classification test performance on CIFAR-100. Except for the learnable leak factor, both architectures and training procedures are identical. Trained for 200 epoch.}
\begin{adjustbox}{max width=0.5\textwidth}

\begin{tabular}{|l|c|}
\hline
\textbf{Network}       & \textbf{CIFAR-100 Accuracy} \\ \hline
 
S-ResNet38 LIF      & 68.71 \%            \\ \hline
S-ResNet38 Parametric LIF & 64.93 \%            \\ \hline
\end{tabular}
\end{adjustbox}
\end{table}

\begin{table}[]
\centering
\caption{\label{tab.leak} Optimal leak coefficient for ResNet38 in CIFAR-100 obtained through two different methods (A single coefficient shared by all layers). "Hyper-parameter search" uses BOHB algorithm to optimize the parameter. The value for this method corresponds to the mean among the 6 best performing configurations found with its corresponding standard deviation in parenthesis. "Learned through PLIF" learns the value by backpropagation during training, the value corresponds to the result after 70 epochs of training.}
\begin{adjustbox}{max width=0.5\textwidth}

\begin{tabular}{|l|c|}
\hline
\textbf{Method}       & \textbf{Leak coefficient} \\ \hline
 
Hyper-parameter search & 0.889 (\(\pm\) 0.003)            \\ \hline
Learned through PLIF & 0.986             \\ \hline
\end{tabular}
\end{adjustbox}
\end{table}

\begin{table}[]
\centering
\caption{\label{tab.single_plif}Image classification test performance on CIFAR-100. In "S-ResNet38 wider + Boost Single PLIF" one single leak value is learned for all layers. Except for the learnable leak factor, both architectures and training procedures are identical. Trained for 70 epoch.}
\begin{adjustbox}{max width=0.5\textwidth}

\begin{tabular}{|l|c|}
\hline
\textbf{Network}       & \textbf{CIFAR-100 Accuracy} \\ \hline
 
S-ResNet38 wider + Boost LIF      & 73.21 \%            \\ \hline
S-ResNet38 wider + Boost Single PLIF & 72.44 \%            \\ \hline
\end{tabular}
\end{adjustbox}
\end{table}

\begin{table}[]
\centering
\caption{\label{tab.pretrain_dvscifar10}Image classification test performance on DVS CIFAR-10. Pre-train column indicates if the network was trained from scratch or pre-trained with a certain dataset. S-Resnet38 stands for the architecture defined in Section \ref{sec.topology} with \(n=6\) and 16 base filters. Wider architectures use 32 base filters and "boosting" indicates the use of a boosting layer (Section \ref{sec.boosting}). Trained for 70 epochs with learning rate reduction at 50\%, 70\% and 90\% of the training process.}
\begin{adjustbox}{max width=0.5\textwidth}

\begin{tabular}{|l|c|c|}
\hline
\textbf{Network}   & \textbf{Pre-train}    & \textbf{DVS CIFAR-10 Acc} \\ \hline
 
S-ResNet38      & No & 63.3 \%            \\ \hline
S-ResNet38 & CIFAR-100   & 70.4 \%            \\ \hline
S-ResNet38 wider + boosting & No & 65.5 \%            \\ \hline
S-ResNet38 wider + boosting & CIFAR-100 & 69.8 \%            \\ \hline
\end{tabular}
\end{adjustbox}
\end{table}

\begin{table}[]
\centering
\caption{\label{tab.pretrain_cifar10}Image classification test performance on CIFAR-10. Pre-train column indicates if the network was trained from scratch or pre-trained with a certain dataset. S-Resnet38 stands for the architecture defined in Section \ref{sec.topology} with \(n=6\) and 16 base filters. Wider architectures use 32 base filters and "boosting" indicates the use of a boosting layer (Section \ref{sec.boosting}). S-Resnet38 Trained for 200 epochs from scratch and for 100 when fine-tuned.  Wider architectures trained for 70 epochs.}
\begin{adjustbox}{max width=0.5\textwidth}

\begin{tabular}{|l|c|c|}
\hline
\textbf{Network}   & \textbf{Pre-train}    & \textbf{CIFAR-10 Acc} \\ \hline
 
S-ResNet38      & No & 91.97 \%            \\ \hline
S-ResNet38 & CIFAR-100   & 92.44 \%            \\ \hline
S-ResNet38 wider + boosting & No & 93.77 \%            \\ \hline
S-ResNet38 wider + boosting & CIFAR-100 & 93.59 \%            \\ \hline
\end{tabular}
\end{adjustbox}
\end{table}

\begin{table}[]
\centering
\caption{\label{tab.dvs_res}Image classification test performance on DVS-CIFAR10. S-Resnet stands for the architecture defined in Section \ref{sec.topology} with 32 base filters, trained for 70 epochs and with CIFAR100 pre-training.}
\begin{adjustbox}{max width=0.5\textwidth}

\begin{tabular}{|l|c|c|}
\hline
\textbf{Network} & \textbf{Resolution}  & \textbf{CIFAR-10 Accuracy}  \\ \hline
 
S-ResNet38  & 32$\times$32   & 71.80 \%             \\ \hline
S-ResNet38 c32k3s2 &  64$\times$64   & 72.98 \%           \\ \hline
S-ResNet38 c32k5s2 MPk2s2  &  128$\times$128   & 72.51 \% \\ \hline
\end{tabular}
\end{adjustbox}
\end{table}

\section{Results} \label{sec.results}
\subsection{State of the art comparison}

In this section we compare our final results to the current state of the art for image classification in the CIFAR-10, CIFAR-100 and DVS-CIFAR10 datasets.

As noted in \cite{fang2021incorporating}, most previous works train on the training set, evaluate the test set at each step, and then report the highest test accuracy obtained. We consider this approach to be reporting validation accuracy rather than test. In our setup, we evaluate the test set after all the training epochs, without using its value for tuning the training. We also evaluate validation accuracy in the same manner than the previous methods in order to make a fair comparison.

The developed S-ResNet outperforms all previous SNN methods in classification accuracy for the CIFAR-10 and CIFAR-100 datasets (Table \ref{tab.sota}). In the DVS-CIFAR10 dataset, we find that the validation accuracy for the best performing network outperforms ours, but when measuring test score, ours is superior.

Before our work, in the CIFAR-10 and CIFAR-100 datasets, the most accurate network was a conversion method. These new results prove how directly training an SNN can perform better without the need of imitating non-spiking computations.

Moreover, in Table \ref{tab.ann} we compare the performance of our S-ResNet to its non-spiking ANN version. We compare the version with 16 and 32 base filters without boosting. We can see how the performance on the trained SNN is not far from its non-spiking counterpart, demonstrating how improvements in SNN training can push these technologies to comparable levels with conventional deep learning.

\begin{figure}[]
\centering
\includegraphics[width=3.5in]{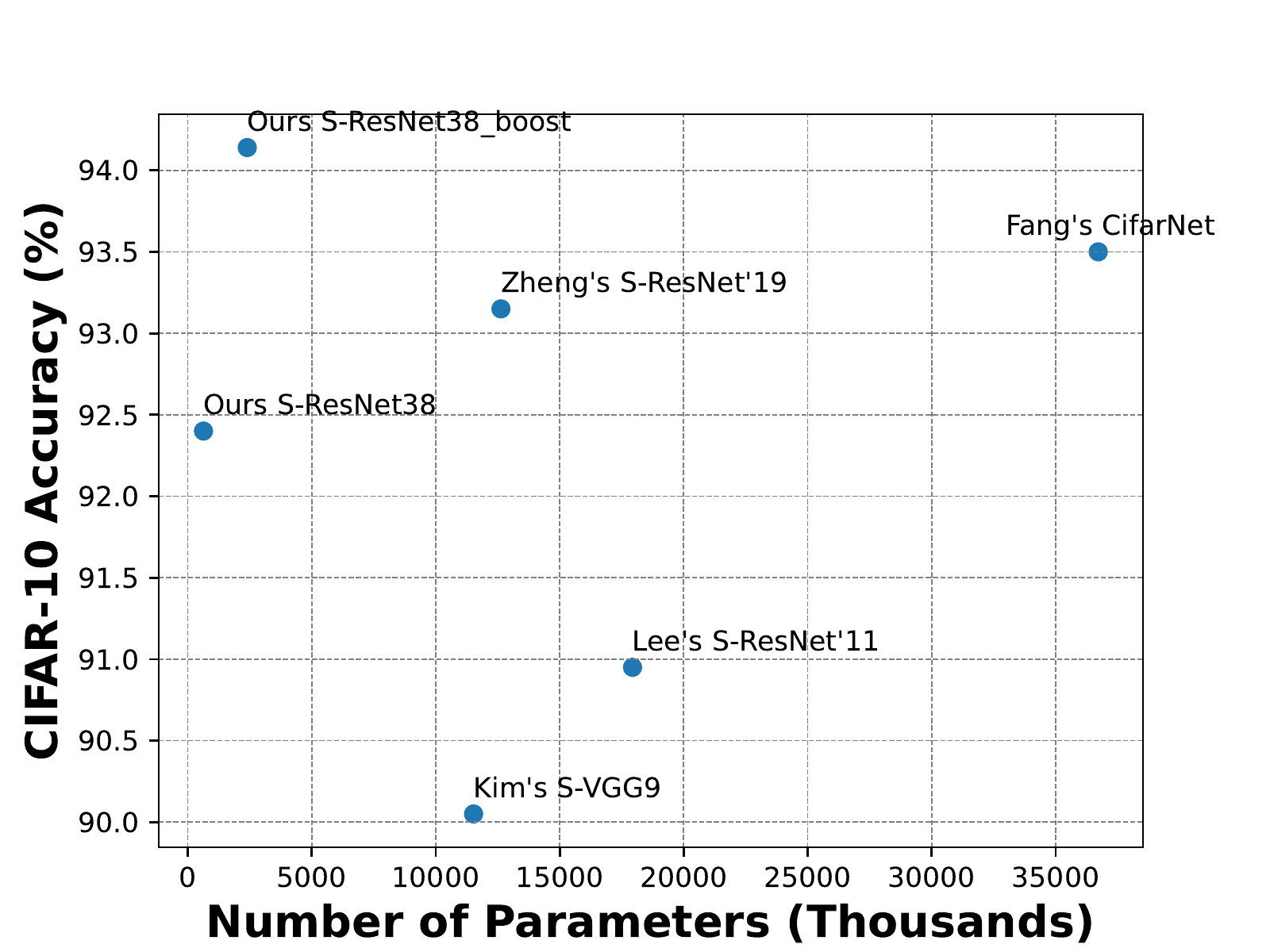}
\caption{CIFAR-10 accuracy versus number of parameters. We compare our network to the best performing trainable SNNs and the other spiking ResNets. "S-ResNet38\_boost" uses the wider architecture with 32 base filters. The number of parameters for other works was counted using their publicly available code. }
\label{fig.cifar10}
\end{figure}

\begin{figure}[]
\centering
\includegraphics[width=3.5in]{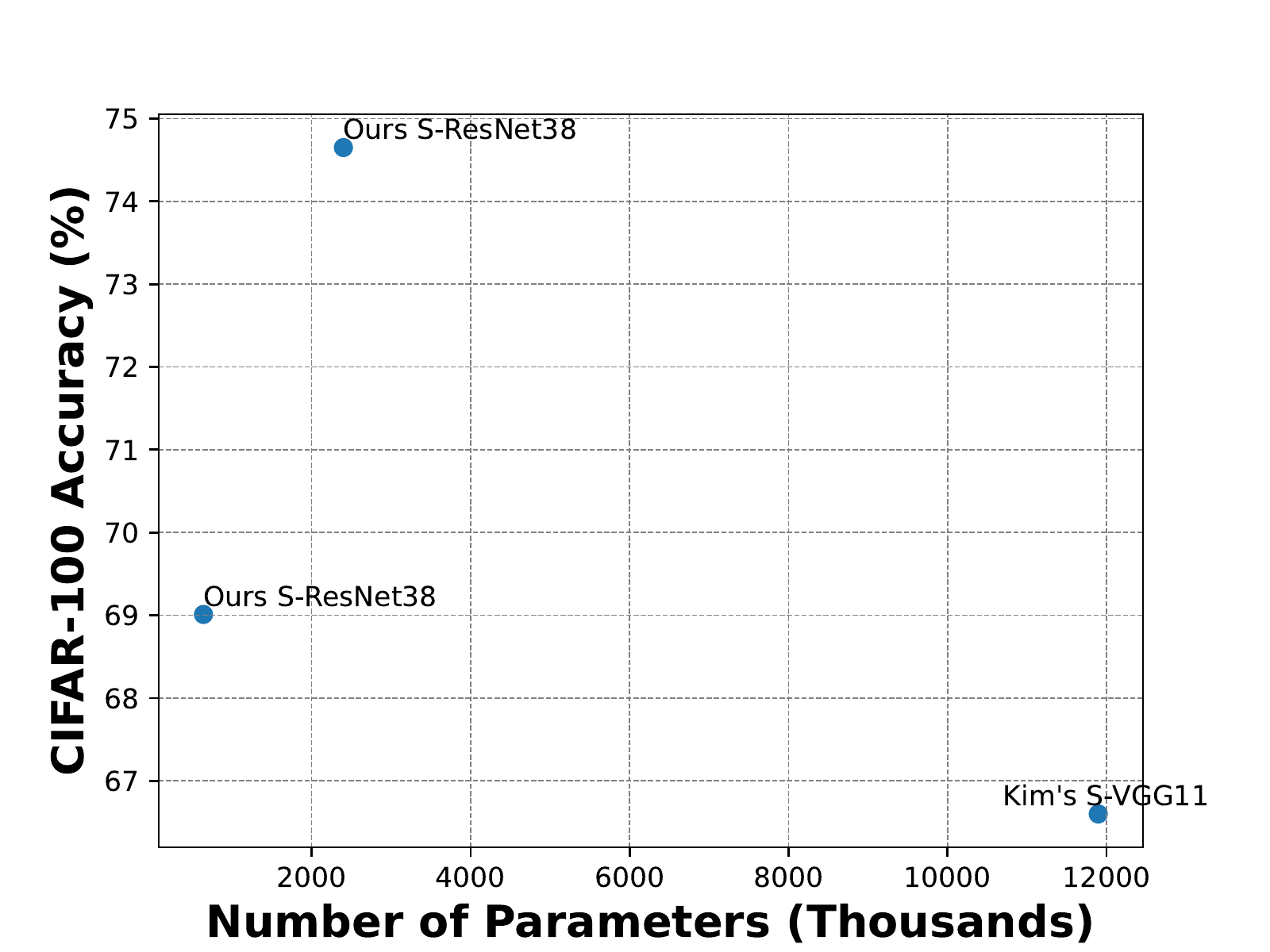}
\caption{ CIFAR-100 accuracy versus number of parameters. We compare our network to the best performing trainable SNN in this dataset. The two results for S-ResNet38 correspond to the same network with 16 or 32 base filters (where 32 base filters has more parameters than 16). The number of parameters for other works was counted using their publicly available code.}
\label{fig.cifar100}
\end{figure}

\begin{figure}[]
\centering
\includegraphics[width=3.5in]{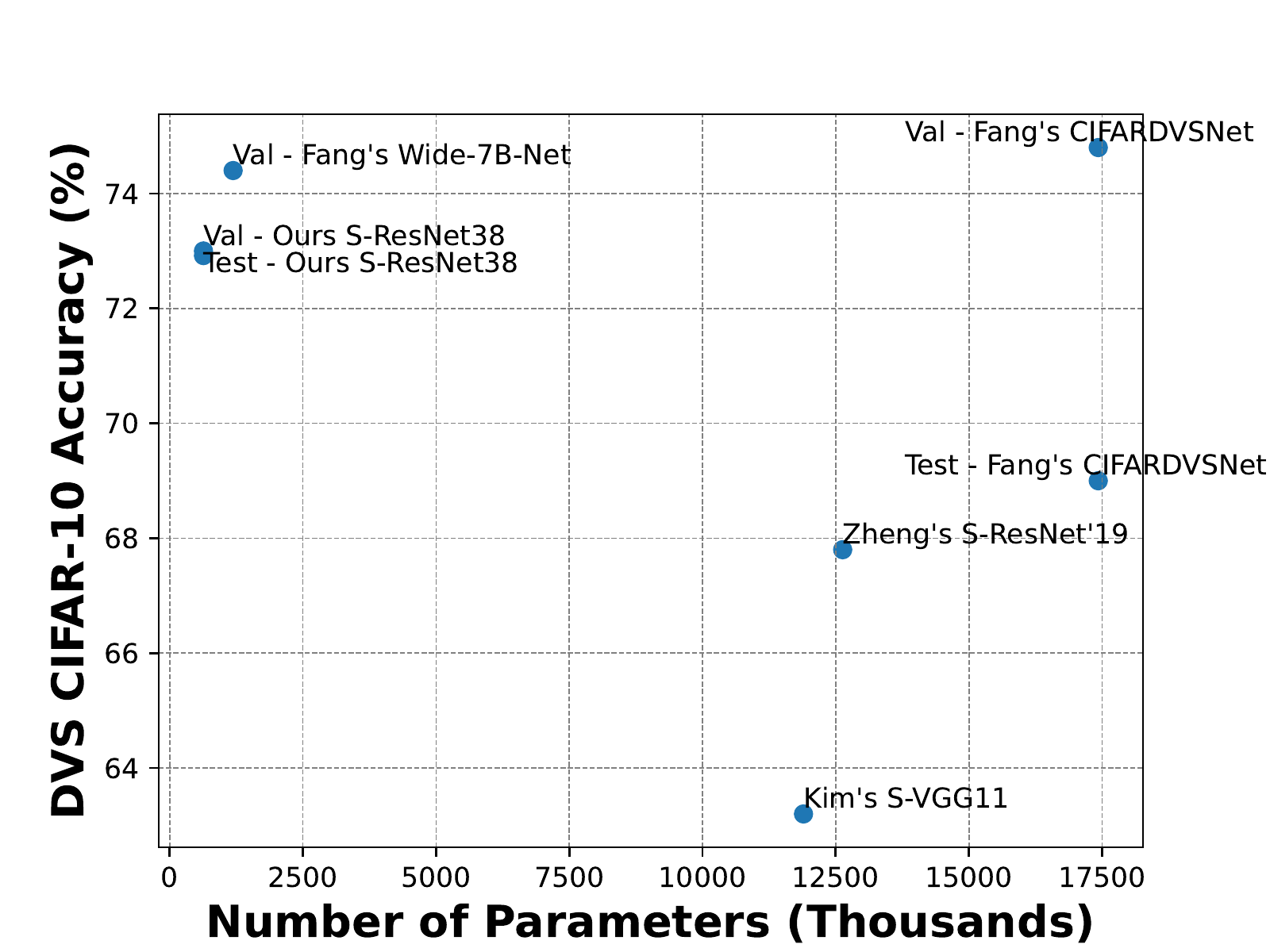}
\caption{DVS CIFAR-10 accuracy versus number of parameters. We compare our network to the best performing trainable SNNs and the other spiking ResNets. The number of parameters for other works was counted using their publicly available code. The "Val" prefix stands for validation accuracy while "Test" stands for testing accuracy. }
\label{fig.dvscifar10}
\end{figure}

Comparing to the previous trainable SNN architectures, our network uses many less parameters. Fig. \ref{fig.cifar10}, \ref{fig.cifar100} and \ref{fig.dvscifar10} show a map of the accuracy versus the number of parameters. The main cause for the difference in parameters is that our network has a smaller number of channels in convolutional layers and only a single fully-connected layer. Then, even when our network is deeper than the others, it is actually lighter in terms of synaptic connections.

\begin{table}[]
\centering
\caption{\label{tab.sota}Image classification validation performance on CIFAR-10, CIFAR-100 and DVS-CIFAR10. Our S-Resnet38 in CIFAR-10 and CIFAR-100 stands for the wider version of the architecture defined in Section \ref{sec.topology} with \(n=6\), 32 base filters, and boosting layer. In DVS-CIFAR10 we use the 16 filters version without boosting and with the pre-training step. We refer to the residual network in \cite{zheng2020going} as S-ResNet', as it follows a different architecture than our S-ResNet }
\begin{adjustbox}{max width=0.5\textwidth}
\centering
\begin{tabular}{|l|l|l|c|}
\hline
\textbf{Network} & \textbf{Method}  & \textbf{Dataset}    & \textbf{Accuracy} \\ \hline
 
Kim \cite{kim2020revisiting} S-VGG9    & Spiking BP  & CIFAR-10 &  90.05 \%            \\
Lee \cite{lee2020enabling} Residual SNN (11)  & Spiking BP & CIFAR-10   &  90.95 \%            \\ 
Zheng \cite{zheng2020going} S-ResNet'19 & Spiking BP & CIFAR-10 &  93.15 \%            \\ 
Fang \cite{fang2021incorporating} CifarNet & Spiking BP & CIFAR-10 &  93.50\%            \\
Wu \cite{wu2021progressive} VGG-11 & SNN conversion & CIFAR-10 &  91.24\% \\
Sengupta \cite{sengupta2019going} VGG-16 & SNN conversion & CIFAR-10 &  91.55\%            \\
Stockl \cite{stockl2021optimized} ResNet-50 & SNN conversion & CIFAR-10 &  92.42\%            \\
Deng \cite{deng2021optimal} ResNet-20 & SNN conversion & CIFAR-10 &  93.58\% \\
 
Han \cite{han2020deep} VGG16 & SNN conversion & CIFAR-10 &  93.63\%            \\ 
 
\textbf{\textit{Ours S-ResNet38}} & \textbf{Spiking BP} & \textbf{CIFAR-10} &  \textbf{94.14\%}            \\ \hline

 Kim \cite{kim2020revisiting} S-VGG9    & Spiking BP  & CIFAR-100 &  66.6 \%            \\ 
 Han \cite{han2020deep} VGG16 & SNN conversion & CIFAR-100 &  70.97\%            \\
 Deng \cite{deng2021optimal} VGG-16 & SNN conversion & CIFAR-100 &  72.34\% \\
 \textbf{\textit{Ours S-ResNet38}} & \textbf{Spiking BP} & \textbf{CIFAR-100} &  \textbf{74.65\%}            \\ \hline 
 
 Kim \cite{kim2020revisiting} S-VGG9    & Spiking BP  & DVS-CIFAR10 &  63.2 \%            \\ 
Zheng \cite{zheng2020going} S-ResNet'19 & Spiking BP & DVS-CIFAR10 &  67.8 \%            \\
Fang \cite{fang2021deep} Wide-7B-Net & Spiking BP & DVS-CIFAR10 &  74.4\%            \\ 
\textbf{Fang \cite{fang2021incorporating} CifarDVSNet} & \textbf{Spiking BP} & \textbf{DVS-CIFAR10} &  \textbf{74.8}\%            \\ 
 \textit{Ours S-ResNet38} & Spiking BP & DVS-CIFAR10 &  72.98\%            \\ \hline
\end{tabular}
\end{adjustbox}

\end{table}

\begin{table}[]
\caption{\label{tab.ann}Image classification performance on CIFAR-10 comparing the ANN version of ResNet to our S-ResNet. All architectures trained for 70 epochs and the same hyper-parameters. S-Resnet38 stands for the architecture defined in Section \ref{sec.topology} with \(n=6\) and 16 base filters. Wider architectures use 32 base filters.}
\centering
\begin{adjustbox}{max width=0.5\textwidth}

\begin{tabular}{|l|c|c|}
\hline
\textbf{Network}   & \textbf{Method}    & \textbf{CIFAR-10 Accuracy} \\ \hline
ResNet38      & ANN & 92.33 \%            \\ \hline
ResNet38 wider     & ANN & 93.56 \%            \\ \hline
S-ResNet38      & SNN & 91.97 \%            \\ \hline
S-ResNet38  wider    & SNN & 92.66 \%            \\ \hline
\end{tabular}
\end{adjustbox}
\end{table}

\subsection{The latency - accuracy compromise}\label{sec.timesteps}
Apart from raw accuracy, the efficiency of algorithms is a major factor when deploying systems in the real world. For image classification in SNN, the amount of time-steps used for prediction regulates a trade-off between accuracy and time or volume of computations.

In order to elucidate the effect of this trade-off in our system, in Table \ref{tab.timesteps} we present the accuracy of S-ResNet38 with different numbers of time-steps. Starting from our best network trained with 50 time-steps, we test how the accuracy degrades when dropping the last 10/20/30/40 steps. Additionally, we compare this to the result obtained by directly training with less time-steps.

The results show how for CIFAR-100, the network trained with 20 steps performs better than dropping the last 30 steps of a 50-step network. Still this same experiment in the CIFAR-10 dataset shows the opposite results by a close margin, indicating that the 50-step network had a more complete training.

At 10 steps, the degrading of the 50-step network becomes more obvious. Interestingly the network trained with 20 time-steps does not degrade as much, as it is only losing half of its computations and therefore still managing to extract the core visual features.

Finally, we hypothesise that the optimal leakage coefficient for the neurons might be correlated to the amount of time-steps the network is ran for. Given that the leak factor that we use was obtained through the hyper-parameter search process, and given that this process prioritized large amounts of time-steps, we believe the optimal leak factor for 20-step inferences could be different from the one we are using. We empirically test this by training the network again with PLIF neurons, a process that allows us to optimize the leak value in a single training run. The results, as seen in Table \ref{tab.20_timesteps}, prove how we obtain a better performance when the leak coefficient is optimized for the number of inference steps, confirming our hypothesis.

From this study we learn how the optimal solution is to perform training with the same amount of time-steps that we want to target at inference time and to optimize hyper-parameters such as the leak factor for this same objective. Still, our SNNs can withstand the effect of early stopping, retaining most of their accuracy even when big percentages of their computation steps are dropped. This allows to provide early estimates in time sensible tasks or to reduce computational cost.

\begin{table}[]
\caption{\label{tab.timesteps} Influence of the number of time-steps in the validation accuracy. Results of the evaluation of the best performing S-ResNet38 with boosting. Training time-steps specifies the number of steps used during training, inference time-steps the steps used for inference. If the inference number is smaller than the training one, early stopping is applied and the last \(N\) time-steps (and learned BNTT layers) are not used. For comparison, the training is reproduced also with 20 time-steps. Clarification: The architecture is the same but the results for CIFAR-100 use the weights trained in CIFAR-100 and the CIFAR-10 results use the weights trained in CIFAR-10.}
\centering
\begin{adjustbox}{max width=0.5\textwidth}
\begin{tabular}{|c|c|c|c|}
\hline
\textbf{Inference t-steps} & \textbf{Training t-steps} & \textbf{CIFAR-100 Acc} & \textbf{CIFAR-10 Acc}\\ \hline
 
50 & 50 &   73.40 \% &   94.10 \%    \\ \hline
40 & 50 &   73.14 \% &   93.96 \%    \\ \hline
30 & 50 &   71.75 \% &   93.61 \%    \\ \hline
20 & 50 &   65.78 \% &   91.93 \%    \\ \hline
20 & 20 &   67.70 \% &   91.28 \%    \\ \hline
10 & 50 &   15.15 \% &   63.86 \%    \\ \hline
10 & 20 &   62.28 \% &   90.45 \%    \\ \hline

\end{tabular}
\end{adjustbox}

\end{table}

\begin{table}[]
\caption{\label{tab.20_timesteps} CIFAR-10 validation accuracy for inferences of 20 time-steps. The first network was trained with 50 time-steps in training time, the others were trained with 20 time-steps. The first two networks use the leak value learned through hyper-parameter optimization done for the 50-step network. The third one optimizes the leak value during its training through PLIF neurons.}
\centering
\begin{adjustbox}{max width=0.5\textwidth}

\begin{tabular}{|c|c|c|c|}
\hline
\textbf{Leak factor} & \textbf{Inference t-steps} & \textbf{Training t-steps} & \textbf{CIFAR-10 Acc}\\ \hline
0.874 & 20 & 50 &   91.93 \%    \\ \hline
0.874 & 20 & 20 &   91.28 \%    \\ \hline
0.995 & 20 & 20 &   92.8 \%    \\ \hline

\end{tabular}
\end{adjustbox}
\end{table}

\section{Conclusions}

In this paper we presented a new SNN architecture which outperforms the previous state of the art in different image classification datasets. This system is the product of an in-depth study on spiking residual connections and design choices based on the empirical results from our experiments. These experiments demonstrate the effects of multiple design choices in the final performance. On top of that, the analysis performed on residual connections sheds new light on the effects of these connections in terms of network activity and hardware requirements. The lessons learned from these studies also become a guide for SNN design, as they allow to make informed choices when building a new SNN feature extractor.

The results of this work demonstrate how SNNs do not need to use conversion methods in order to maximize their accuracy. Additionally, they contribute to pushing their performance closer to that of non-spiking deep learning. From here, we hope that new applications can benefit from increased accuracy by fine tuning our networks and more experiments can follow in order to keep pushing the SNN state of the art.

\section*{Acknowledgement}
This work was supported by the US Air Force Office of Scientific Research under Grant for project FA8655-20-1-7037. The contents were approved for public release under case AFRL-2021-3889 and they represent the views of only the authors and does not represent any views or positions of the Air Force Research Laboratory, US Department of Defense, or US Government. The authors declare that there is no conflict of interest.

{\small
\bibliographystyle{unsrt}
\bibliography{bib_collection_14-10-21}
}
\end{document}